\documentclass[sigconf,authorversion]{acmart}

\usepackage[labelformat=simple]{subcaption}
\usepackage{tabularx}

\usepackage{booktabs}  

\usepackage{graphicx}
\usepackage{bm}
\usepackage{url}
\usepackage{balance}

\usepackage{bbm}  
\usepackage{multirow}
\usepackage{algorithm}
\usepackage{algpseudocode}

\usepackage{hyperref}

\renewcommand{\vec}[1]{\mathbf{#1}}

\newcommand{\vmsg}{\mathbf{m}}
\newcommand{\vemb}{\mathbf{h}}
\newcommand{\vedge}{\mathbf{e}}
\newcommand{\vmsgmem}{\mathbf{s}}
\newcommand{\vupdmem}{\mathbf{h}'}
\newcommand{\vgnn}{\mathbf{z}}
\newcommand{\vnode}{\mathbf{v}}

\newcommand{\mem}{\mathcal{M}}
\newcommand{\hist}{\mathcal{H}}

\DeclareMathOperator{\msg}{\mathtt{msg}}

\DeclareMathOperator{\upd}{\mathtt{upd}}
\DeclareMathOperator{\merge}{\mathtt{mlp}}
\DeclareMathOperator{\mha}{\mathtt{mha}}
\DeclareMathOperator{\transformer}{\mathtt{transformer}}
\DeclareMathOperator{\reindex}{\mathtt{reindex\_emb}}

\newcommand{\ie}[0]{\textit{i.e.}}
\newcommand{\eg}[0]{\textit{e.g.}}

\newcommand{\tpm}[0]{$\pm~$}

\copyrightyear{2023}
\acmYear{2023}
\setcopyright{acmlicensed}

\acmConference[WWW '23]{Proceedings of the ACM Web Conference 2023}{May 1--5, 2023}{Austin, TX, USA}
\acmBooktitle{Proceedings of the ACM Web Conference 2023 (WWW '23), May 1--5, 2023, Austin, TX, USA}
\acmPrice{15.00}
\acmDOI{10.1145/3543507.3583433}
\acmISBN{978-1-4503-9416-1/23/04}

\newcommand{\ours}[0]{TIGER}
\newcommand{\oursfull}[0]{\textbf{T}emporal \textbf{I}nteraction \textbf{G}raph \textbf{E}mbedding with \textbf{R}estarts}

\begin{document}

\title{TIGER: Temporal Interaction Graph Embedding with Restarts}

\author{Yao Zhang}
\email{yaozhang@fudan.edu.cn}
\author{Yun Xiong}
\authornote{Corresponding author.}
\email{yunx@fudan.edu.cn}
\affiliation{%
  \institution{Shanghai Key Laboratory of Data Science, School of Computer Science, Fudan University}
  \country{China}
}

\author{Yongxiang Liao}
\email{liaoyx21@m.fudan.edu.cn}
\affiliation{%
  \institution{Shanghai Key Laboratory of Data Science, School of Computer Science, Fudan University}
  \country{China}
}

\author{Yiheng Sun}
\email{elisun@tencent.com}
\affiliation{%
  \institution{Tencent Weixin Group}
  \country{China}
}

\author{Yucheng Jin}
\email{ycjin22@m.fudan.edu.cn}
\affiliation{%
  \institution{Shanghai Key Laboratory of Data Science, School of Computer Science, Fudan University}
  \country{China}
}

\author{Xuehao Zheng}
\email{xuehaozheng@tencent.com}
\affiliation{%
  \institution{Tencent Weixin Group}
  \country{China}
}

\author{Yangyong Zhu}
\email{yyzhu@fudan.edu.cn}
\affiliation{%
  \institution{Shanghai Key Laboratory of Data Science, School of Computer Science, Fudan University}
  \country{China}
}

\renewcommand{\shortauthors}{Zhang et al.}

\begin{abstract}
Temporal interaction graphs (TIGs), consisting of sequences of timestamped interaction events, are prevalent in fields like e-commerce and social networks. To better learn dynamic node embeddings that vary over time, researchers have proposed a series of temporal graph neural networks for TIGs. However, due to the entangled temporal and structural dependencies, existing methods have to process the sequence of events chronologically and consecutively to ensure node representations are up-to-date. This prevents existing models from parallelization and reduces their flexibility in industrial applications. To tackle the above challenge, in this paper, we propose TIGER, a TIG embedding model that can restart at any timestamp. We introduce a restarter module that generates surrogate representations acting as the warm initialization of node representations. By restarting from multiple timestamps simultaneously, we divide the sequence into multiple chunks and naturally enable the parallelization of the model. Moreover, in contrast to previous models that utilize a single memory unit, we introduce a dual memory module to better exploit neighborhood information and alleviate the staleness problem. Extensive experiments on four public datasets and one industrial dataset are conducted, and the results verify both the effectiveness and the efficiency of our work.
\end{abstract}



\maketitle

\section{Introduction}
\label{sec:intro}

\begin{figure}
\begin{tabular}{@{}c@{}}
\begin{subfigure}{\linewidth}
\centering
\includegraphics[width=.8\linewidth]{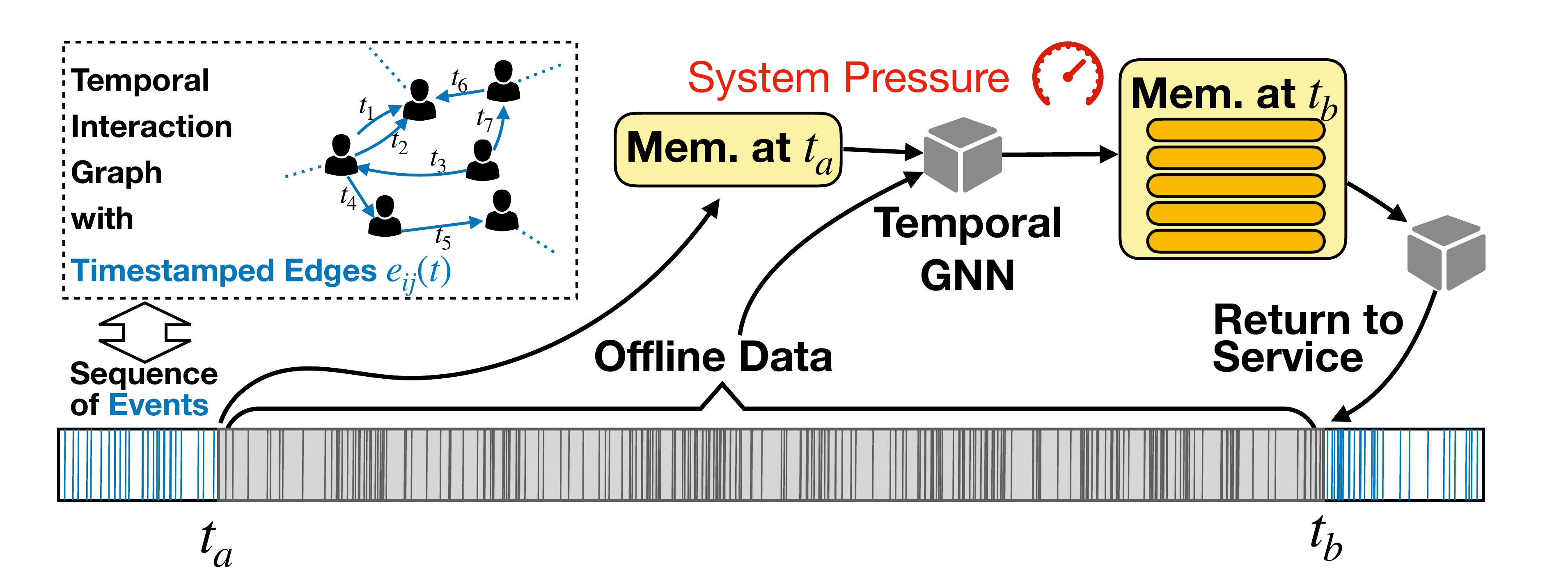} 
\caption{Previous temporal GNNs \cite{jodie,deepcoevolve,rossi2020temporal} without restarter}
\label{fig:butterfly:tgn}
\end{subfigure}
\\
\begin{subfigure}{\linewidth}
\centering
\includegraphics[width=.8\linewidth]{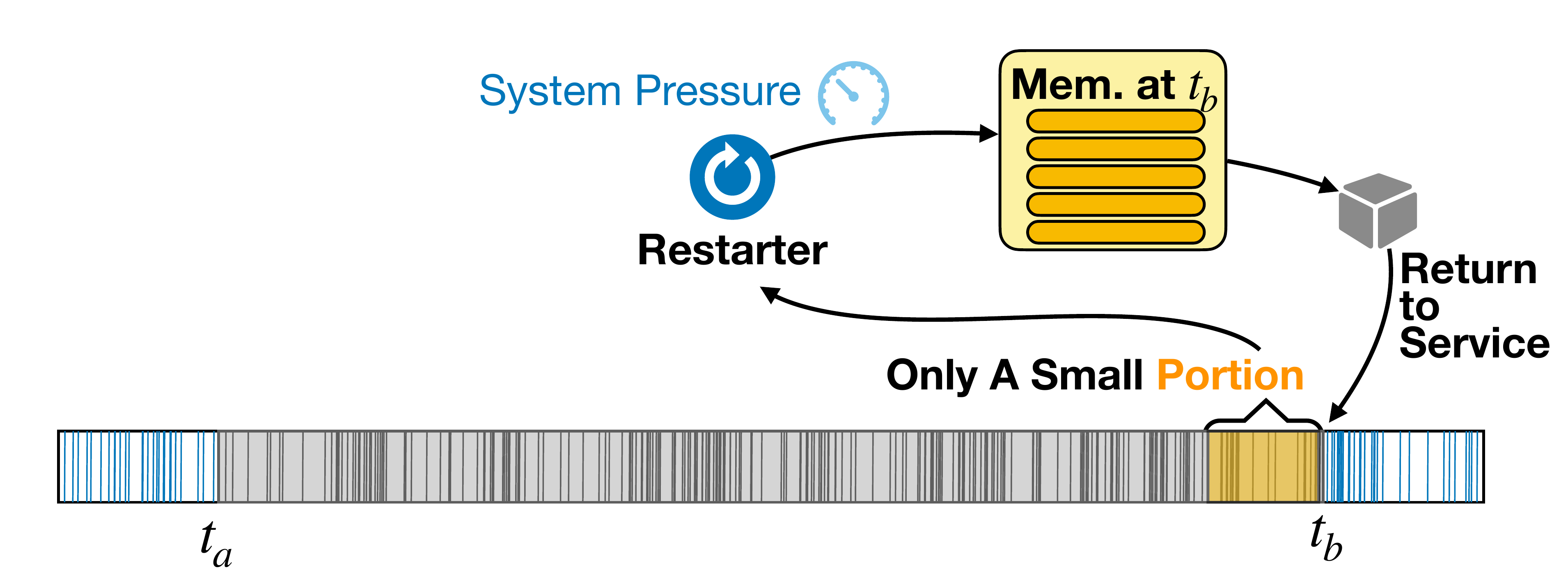} 
\caption{Temporal GNNs with restarter [This paper]}
\label{fig:butterfly:restart}
\end{subfigure}
\\
\begin{subfigure}{\linewidth}
\centering
\includegraphics[width=.8\linewidth]{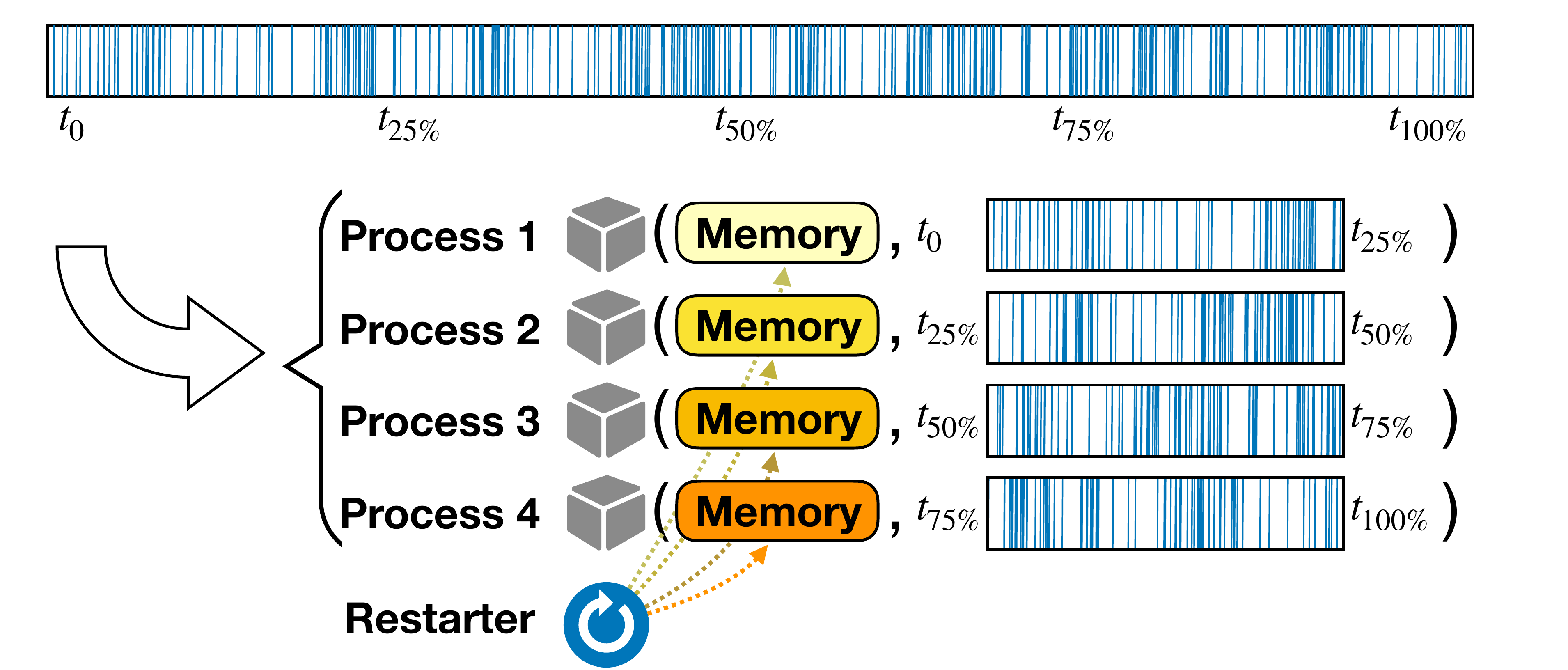} 
\caption{Parallelization of \ours{} [This paper]}
\label{fig:butterfly:parallel}
\end{subfigure}
\end{tabular}
\caption{%
Comparison of methods with and without restarters.
(a) The previous methods need to consume the offline data to compute the up-to-date memory before returning to service.
(b) With the help of the restarter, our proposed methods can quickly return to service with the estimated memory.
(c) The restarter also enables our proposed methods to run in parallel.
}
\label{fig:butterfly}
\Description{Comparison of methods with and without restarters.}
\end{figure}

\begin{figure*}[!t]
\begin{tabular}{@{}c@{}c@{}}
\begin{subfigure}{.39\linewidth}
\includegraphics[width=\linewidth]{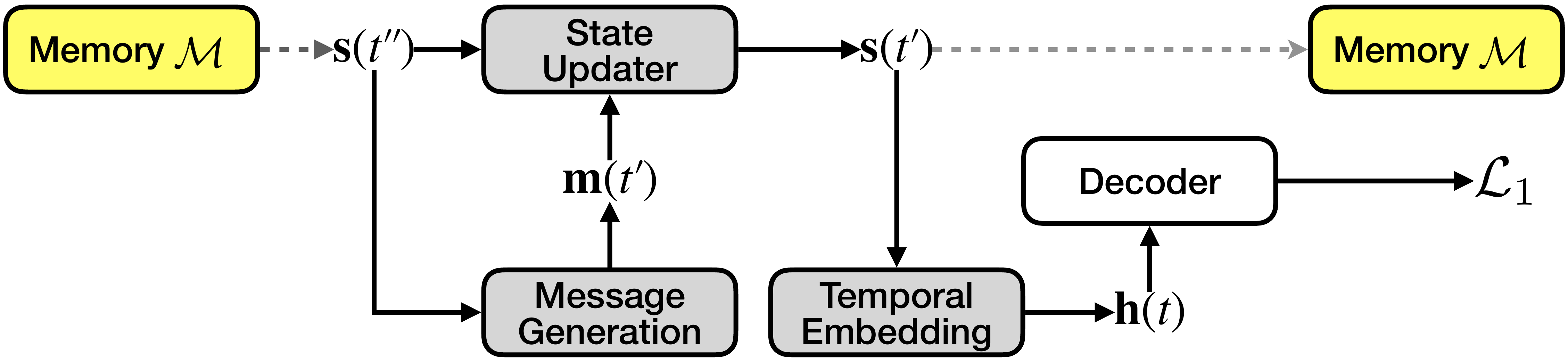}
\caption{TGN \cite{rossi2020temporal}}
\label{fig:dataflow:tgn}
\end{subfigure}
&
\begin{subfigure}{.39\linewidth}
\includegraphics[width=\linewidth]{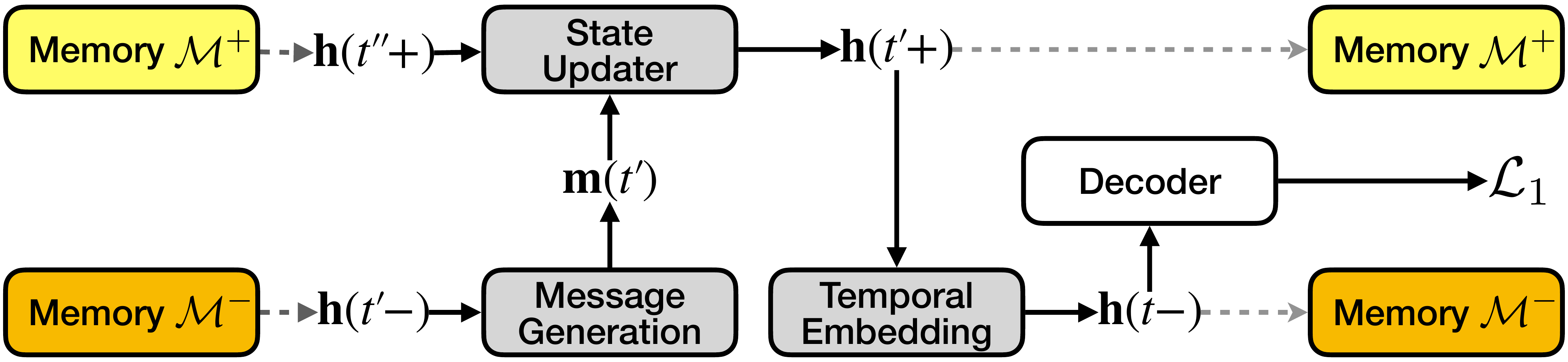}
\caption{TIGE [This paper]}
\label{fig:dataflow:tige}
\end{subfigure}
\\
\multicolumn{2}{c}{
\begin{subfigure}{.78\linewidth}
\includegraphics[width=\linewidth]{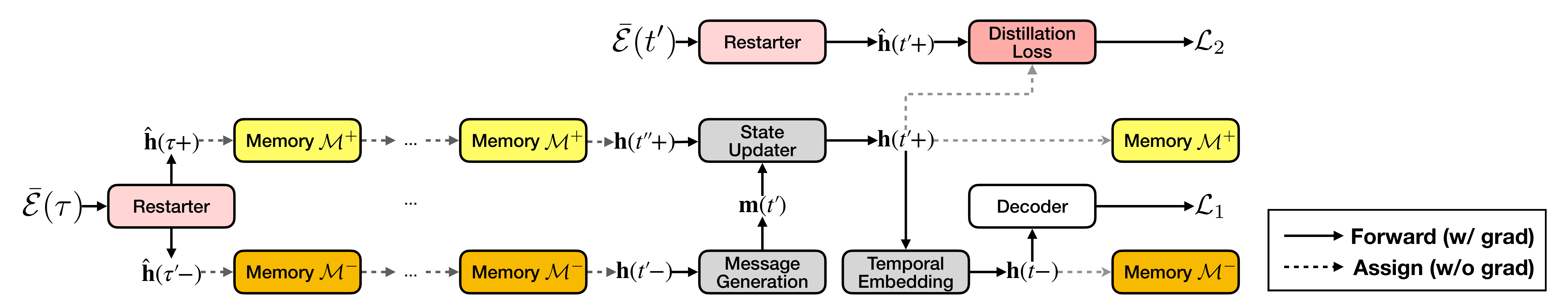}
\caption{TIGER [This paper]}
\label{fig:dataflow:tiger}
\end{subfigure} 
}
\end{tabular}
\caption{%
Illustration of data flows during training. 
Note that to make gradients pass through state updaters and message generators, we need to use penultimate events at time $t''$.
(a) TGN contains only one memory unit $\mathcal{M}$, and the output of the temporal embedding module is discarded after computing the loss $\mathcal{L}_1$.
(b) TIGE introduces the dual memory to track node representations before and after events.
Compared to TGN, our proposed TIGE has explicit meanings of values in memory and takes full use of the outputs of the temporal embedding module.
(c) In TIGER, the restarter occasionally re-initializes the memory during training.
The training target of the restarter is to mimic the outputs of the encoder. ($\hat{\vemb}(t'-)$ is not shown here for simplicity.)
}
\label{fig:dataflow}
\Description{Illustration of data flows during training}
\end{figure*}

Many real-world systems can be formulated as temporal interaction graphs (TIGs) comprising sequences of timestamped interaction events, \ie, edges, among objects, \ie, nodes.
Different from static graph representation learning \cite{velivckovic2017graph}, temporal interaction graph embedding aims to encode temporal and structural dependencies at the same time and produce node representations suitable for various downstream tasks \cite{jodie,xu2020inductive,rossi2020temporal,zhang2021cope}.

Early attempts at temporal interaction graph embedding mainly focus on generating static representations \cite{ige,igep,tigecmn,ctdne}.
However, these methods cannot handle new nodes or edges as graphs evolve.
On the other hand, static representations are not enough for time-aware prediction tasks in real scenarios.
For example, we may predict whether a user is fraudulent given his/her recent transactions in an online trading network, or whether a user, with evolving desires, will purchase an item in the future in recommender systems.

To capture the dynamic nature of TIGs and learn time-aware representations, researchers have proposed temporal graph neural networks (GNNs) that can jointly model temporal and structural dependencies and generalize to new nodes and edges \cite{deepcoevolve,jodie,xu2020inductive,rossi2020temporal,zhang2021cope}.
As a critical element of most temporal GNNs \cite{jodie,rossi2020temporal}, the memory module stores node states representing the past or historical information of nodes.
This enables the temporal GNNs to memorize long-term dependencies and thus achieve superior performance than memory-less models \cite{xu2020inductive,rossi2020temporal}.

Despite their success, these models are subject to two significant problems.
Due to the \textbf{complex dependency} of temporal interaction graphs, the previous methods cannot be easily parallelized.
Unlike static graphs, in temporal GNNs, the message passing procedures must conform to the temporal constraints: a node cannot aggregate information from future neighbors.
On the other hand, unlike personalized sequence models that can process multiple independent sequences in parallel, in temporal GNNs, all nodes' events are entangled.
The combination of the structural and temporal dependencies requires the models to traverse the past edges chronologically and consecutively so that the memory can stay up-to-date.

This severely reduces the flexibility of applications in industrial scenarios.
Consider a case in an online payment platform where a deployed model was suspended on $t_a$, the first day of this month, due to, for example, data center migration, and at $t_b$, the end of the month, we want it back in service.
Then we need to process billions of transactions that happened during the model offline time as illustrated in Figure \ref{fig:butterfly:tgn}.

Another drawback of the previous methods is the so-called \textbf{``staleness'' problem} \cite{kazemi2020representation}.
Since the memory of a node only gets updated when events involving the node happen, its state would become stale if it is inactive for a while, leading to piecewise constant temporal embeddings.
To deal with this problem, researchers have proposed temporal embedding modules \cite{jodie,rossi2020temporal,zhang2021cope}.
For example, TGN \cite{rossi2020temporal}, as one of the most representative temporal GNNs, introduces temporal graph attention layers \cite{xu2020inductive} to aggregate information from a node's neighbors since some neighbors may have been updated recently.

However, by analyzing the data flow, we will find that the memory updating schema still suffers from the staleness problem in these methods due to their two-branch structure.
The branch with temporal embedding modules can indeed generate fresh representations.
But the branch in charge of memory updating completely neglects this fresh information, leading to stale memory.

In this paper, we tackle the above problems by proposing a new temporal GNN framework \ours{} (\oursfull).
In TIGER, we introduce a new module called restarter.
The restarter is trained to mimic the memory and could infer the memory state at any time with a negligible cost. 
The outputs of the restarter can then be used as warm initialization of the memory so that we can restart the model flexibly at any time as illustrated in Figure \ref{fig:butterfly:restart}.
Another benefit of introducing restarter is that we can run our model in parallel by dividing the sequence of events into multiple chunks as shown in Figure \ref{fig:butterfly:parallel},
which further improves the flexibility and scalability of our methods.

To deal with the staleness problem, we propose a dual memory module.
The dual memory module unifies the memory updating branch and the temporal embedding branch so that the fresh representations produced by the temporal embedding module are fully utilized and feedback the memory as shown in Figure \ref{fig:dataflow}.
Moreover, we explicitly distinguish between the node representations just before the events and right after the events and store them in the corresponding memory units.
This makes our memory module more interpretable than previous methods \cite{rossi2020temporal}.

The main contributions of this paper are as follows.
(1) We propose a dual memory-based temporal graph neural network that further alleviates the staleness problem by merging the memory update branch and the temporal embedding branch. Moreover, by differentiating node representations before and after events, our proposed methods are more interpretable.
(2) We propose the restarter module, whose training objective is to directly mimic the node representations so that it can re-initialize the memory warmly. This enables our models to restart at any time and run in parallel.
(3)  We conduct extensive experiments on four public datasets and one industrial dataset.
Notably, our models with only 20\% of training data can outperform strong baselines trained with complete data,
and the multi-GPU parallelization experiment shows satisfying speedups.
These experimental results prove the effectiveness and efficiency of the proposed models.

\section{Related Work}
\label{sec:related}

Early works \cite{htne,ctdne,ige,igep,tigecmn} mainly focus on learning static representations of temporal interaction graphs.
For example, CTDNE \cite{ctdne} generates node embeddings by simulating temporal random walks.
IGE and its variants \cite{ige,igep,tigecmn} model temporal dependency by extracting node-induced sequences with coupled prediction networks.
However, these methods cannot generalize to new nodes and edges and fail to depict the evolving nature of TIGs.
Instead of learning static node embeddings, DeepCoevolve \cite{deepcoevolve} generates dynamic node representations by a pair of recurrent networks.
However, the method generates piecewise constant representations and suffers from the staleness problem \cite{kazemi2020representation}.
JODIE \cite{jodie} extends the idea and introduces the projection operation to make representations more time-aware.
DyRep \cite{trivedi2019dyrep} uses attention layers to aggregate neighbor information.
The above methods need to maintain the dynamic states of nodes, which are then unified as memory modules in TGN \cite{rossi2020temporal}.
In contrast to the above methods, TGAT \cite{xu2020inductive} is a memory-less method utilizing temporal attention layers equipped with functional time embedding \cite{xu2019self}.
TGN \cite{rossi2020temporal} generalizes the previous methods by differentiating the model into the memory module, the message-related modules, and the embedding module and achieves state-of-the-art performance.

Only a few studies on the efficiency of temporal GNNs have been made.
APAN \cite{wang2021apan} introduces a mailbox storing neighbor states for each node so that the temporal graph aggregation operations can be done locally during the online inference.
But it only accelerates the inference speed by space-time tradeoff.
TGL \cite{zhou2022tgl} is a library implementing several existing temporal GNNs like TGN.
It focuses on optimizing GPU-CPU communication.
EDGE \cite{chen2021efficient} decouples edges by detecting d-nodes in the graph with integer programming and assigning static embeddings to d-nodes.
However, it is only applicable to JODIE, which has been proven inferior to other temporal GNNs \cite{rossi2020temporal,xu2020inductive}.
These approaches are orthogonal to our work as we focus on how we can restart the model at any time with the warmly re-initialized memory.

There are also some recent works on TIGs but beyond the scope of our paper.
For example, besides the above studies on self-supervised representation learning of TIGs, the researchers proposed methods like CAW-N \cite{wang2021inductive} and NAT \cite{luo2022neighborhood} that are dedicated to temporal link prediction tasks by examining joint structure information.
Nevertheless, the generated representations are less effective on downstream tasks \cite{tian2021self}.
In \cite{tian2021self}, the authors proposed a novel self-supervised learning framework for TIGs,
and in \cite{wang2021adaptive}, the authors focus on data augmentation on TIGs.

\section{Methodology}
\label{sec:method}

A temporal interaction graph comprises a node set $\mathcal{V} = \{1, \dots, |\mathcal{V}| \}$ and an edge set $\mathcal{E} = \{e_{ij}(t) \}$ where $i, j \in \mathcal{V}$.
The edge set can be viewed as a stream of events, and each edge $e_{ij}(t)$ corresponds to an interaction event between the source node $i$ and the destination node $j$ at timestamp $t \ge 0$.
We denote node features and edge features as $\vnode_i(t)$ and $\vedge_{ij}(t)$ respectively.
If the graph is non-attributed, we simply assume $\vnode_{\cdot}(t) = \vec{0}$ and $\vedge_{\cdot}(t) = \vec{0}$, \ie, zero vectors.
For the sake of simplicity, we assume the dimensions of node/edge features and node representations are $d$ throughout the paper.
We use $\mathcal{E}(t) = \{ e_{ij}(\tau) \in \mathcal{E}| \tau < t \} $ to denote the set of events before time $t$ and  use $\hist_i(t) = \{ e_{ij}(\tau)|\tau < t \} \cup \{ e_{ji}(\tau)|\tau < t \}$  to denote the set of events involving node $i$ before time $t$, which can be viewed as the history of node $i$.
Moreover, we use $\bar{\hist}_i^m(t)$ as the truncated history of node $i$ where only the most recent $m$ events are kept.

\subsection{Overview}

In this paper, we follow the setting in \cite{kazemi2020representation} and describe our proposed \ours{} from an encoder-decoder view.
The encoder takes a temporal graph as input and produces temporal representations of nodes.
As a critical ingredient of the encoder, a dual memory module is introduced to track the evolving node representations and memorize long-term dependencies.
The decoder then utilizes the node representations and performs a series of downstream tasks.
Here, we consider the temporal link prediction task so as to provide self-supervision signals to train the model.
The encoder and the decoder constitute a base model named TIGE.
To further increase the scalability and flexibility, \ours{} contains an extra module, \ie, the restarter. 
At any timestamp, the restarter can re-initialize the memory in the encoder using only a small portion of the events.
This enables our model to restart the training/inference process at any time, and thus run in parallel.

\subsection{Encoder}
\label{sec:method:enc}

The encoder of our proposed \ours{} contains four components.
A dual memory module is introduced to track the evolving node representations.
By fetching node representations from the memory, the message generator can generate messages about new-arriving events.
Then the state updater consumes the messages and updates node representations to reflect the impacts of the events.
Considering the state updater only work when events happen, which leads to the staleness problem \cite{kazemi2020representation}, the temporal embedding module utilizes temporal graph attention layers to dynamically generate pre-jump node representations at any time.

To make expressions less cumbersome, we introduce these components relatively independently.
The complete data flow among them is illustrated in Figure \ref{fig:dataflow}.

\subsubsection{Dual Memory}

Given an event $e_{ij}(t)$ between node $i$ and node $j$ at time $t$, in this paper, we assume that node $i$ is associated with two representations at this moment (so does node $j$):
$\vemb_i(t-)$, the representation just before the event,
and $\vemb_i(t+)$, the representation right after the event.
The former can be used to predict if the event will happen, while the latter reflects the impact of the event.
From this view, if we treat the temporal embedding $h_i$ as a function of $t$, then $\{ t=\tau | \exists e_{i\cdot}(\tau) \in \mathcal{E} \text{ or } e_{\cdot i}(\tau) \in \mathcal{E} \}$ are the jump discontinuities of the function. 
We call $\vemb(t-)$ and $\vemb(t+)$ the pre- and post-jump representations respectively.

In this paper, we introduce two memory units, $\mathcal{M}^+$ and $\mathcal{M}^-$, which store representations $\vemb_i(t+)$ and $\vemb_i(t-)$ for each node $i$.
For each node, its memory values $\mem_i^+$ and $\mem_i^-$ are first initialized as zero vectors and then get updated over time as new events involving the node arrive.

Most previous methods \cite{jodie,rossi2020temporal} only utilize one single memory without explicit meaning as shown in Figure \ref{fig:dataflow:tgn}.
In the following Equation \ref{eq:msg_gen} and Equation \ref{eq:upd} we will see how dual memory can help generate better representations.

\subsubsection{Message Generation}

For an event at time $t$ involving node $i$, a message $\vmsg_i(t)$ is first generated to compute the node representation after the event, \ie, $\vemb(t+)$.

Assume the event $e_{ij}(t)$ is between node $i$ and node $j$, with the feature vector $\vedge_{ij}(t)$.
Firstly, we fetch from the memory the previous states of the nodes:
$\vmsgmem_i(t)  \leftarrow \mathcal{M}^{\text{msg}}_i$ and $ \vmsgmem_j(t) \leftarrow \mathcal{M}^{\text{msg}}_j$,
where $\mem^{\text{msg}} \in \{\mem^+, \mem^- \}$.
If we fetch values from $\mathcal{M}^+$, we have $\vmsgmem_i(t) = \vemb_i(t_i'+)$, which is the state of node $i$ right after its last event at time $t_i'$.
Another option is to use $\mathcal{M}^-$, and we then have $\vmsgmem_i(t)=\vemb_i(t-)$ representing the state of node $i$ just before the current event.
By default we use $\mem^{\text{msg}} = \mem^-$ as it not only considers the impact of the last event at $t'$, but also captures the evolution of nodes within the time window $(t', t)$.
After introducing the temporal embedding in Section \ref{sec:method:temb}, we can notice that $\vemb_i(t-)$ also encodes the temporal neighborhood information, which makes it more expressive than $\vemb_i(t'+)$.

Two messages are then generated by
\begin{equation}
\begin{aligned}
{\vmsg}_i(t) &= \msg\big(\vmsgmem_i(t), \vmsgmem_j(t), \vedge_{ij}(t), \Phi(t-t_i')\big), \\
{\vmsg}_j(t) &= \msg\big(\vmsgmem_j(t), \vmsgmem_i(t), \vedge_{ij}(t), \Phi(t-t_j')\big), \\
\end{aligned}
\label{eq:msg_gen}
\end{equation}
where $\msg$ is the message function, and $t_j'$ is the timestamp of the last event involving node $j$.
The time encoding function $\Phi$ \cite{xu2019self,xu2020inductive} maps the time interval into a $d$-dimension vector.
For simplicity, we use the commonly-used \textit{identity} message function that returns the concatenation of the input vectors.
Beside, we process multiple consecutive events in a batch for the sake of efficiency.
We use the simplest \textit{most recent message} aggregator that only keeps the most recent message for each node in the batch as \cite{rossi2020temporal}.

\subsubsection{State Updater}

As we mentioned earlier, the state of nodes $\vemb(t+)$ reflects the impacts of events.
We compute the new representation for node $i$ after receiving the event $e_{ij}(t)$:
\begin{equation}
\begin{aligned}
\vupdmem_i(t) &\leftarrow \mem_i^{\text{upd}},\\
\vemb_i(t+) &= \upd(\vupdmem_i(t), \vmsg_i(t)),
\end{aligned}
\label{eq:upd}
\end{equation}
where $\upd$ is the learnable update function, and we use GRU \cite{cho2014learning} in practice.
Again, there are two sources for the previous state $\vupdmem_i(t)$, \ie, 
$\mem^{\text{upd}} \in \{\mem^+, \mem^- \}$.
If we use $\vupdmem_i(t) = \vemb_i(t-)$, then Equation \ref{eq:upd} can be viewed as aiming to model the instant impact of the event and simulate the jump at time $t$ \cite{zhang2021cope}.
On the other hand, when $\vupdmem_i(t) = \vemb_i(t_i'+)$, Equation \ref{eq:upd} is more like traditional recurrent units where $\vemb(t'+)$ acts as the last hidden state and $\vmsg(t)$ as the input.
By default we use the latter as it is more compatible with the GRU update function.
After computing the new embedding $\vemb_i(t+)$ for node $i$, we update the memory $\mem^+$ by overwriting the corresponding value, 
$\mem^+_i \leftarrow \vemb_i(t+)$.

\subsubsection{Temporal Embedding}
\label{sec:method:temb}

The temporal embedding module aims to generate the pre-jump representations $\vemb(t-)$ at any time $t$ before the next event arrives.
Specifically, we adopt an $L$-layer temporal graph attention network \cite{xu2020inductive} to aggregate neighborhood information, which can alleviate the ``staleness'' problem \cite{kazemi2020representation}.

We first fetch data from memory $\vgnn_k = \vemb_k(t'_k+) \leftarrow \mem^+_k $ for each node $k$ that are involved in the $L$-hop receptive fields of node $i$ and augment them with their node features:
$\vgnn_k^{(0)} = \vgnn_k + \vnode_k(t)$,
where $\vgnn_k^{(0)}$ is the input of the first temporal graph attention layer.
Note that at this moment, $\mem_k^+$ stores the post-jump representation which was computed after its last event happened at time $t'_k$.

Then for each layer $1 \leq l \leq L$, we use the multi-head attention \cite{vaswani2017attention} to aggregate neighborhood information as
\begin{equation}
\begin{aligned}
\vgnn_k^{(l)} &= \merge^{(l)}(\vgnn_k^{(l-1)} \| \tilde{\vgnn}_k^{(l)}), \\
\tilde{\vgnn}_k^{(l)} &= \mha^{(l)}\big(\vec{q}_k^{(l)}, \vec{K}_k^{(l)}, \vec{V}_k^{(l)}),\\
\vec{q}_k^{(l)} &= \vgnn_k^{(l-1)} \| \Phi(0), \\
\vec{K}_k^{(l)} &= \vec{V}_k^{(l)} = 
\begin{bmatrix}
\vgnn^{(l-1)}_{\pi_k(1)} \| \vedge_{\pi_k(1)}\| \Phi(t - t_{\pi_k(1)}),\\
\dots \\ 
\vgnn^{(l-1)}_{\pi_k(n)} \| \vedge_{\pi_k(n)}\| \Phi(t - t_{\pi_k(n)}),
\end{bmatrix}
\end{aligned}
\label{eq:tgat}
\end{equation}
where $\|$ denotes vector concatenation, $\merge^{(\cdot)}$ are two-layer feed-forward networks (MLPs) with hidden/output dimension being $d$, and $\mha^{(\cdot)}$ are multi-head attention functions with queries $\vec{q}^{(\cdot)} \in \Re^{2d}$, keys $\vec{K}^{(\cdot)} \in \Re^{n\times 3d}$ and values $\vec{V}^{(\cdot)} \in \Re^{n \times 3d}$.
We limit the receptive fields to the most recent $n$ events for each node $k$, denoted by $\{e_{\pi_k(1)}(t_{\pi_k(1)}), \dots,  e_{\pi_k(n)}(t_{\pi_k(n)})\}$.
Here $\pi$ is a permutation, and $\pi_k(\cdot)$ are temporal neighbors of node $k$.
We omit subscript $k$ in $e_{\pi_k(\cdot)}$, which may refer to an event from $k$ to $\pi_k(\cdot)$ or vice versa.

We use $\vemb_i(t-) = z_i^{(L)}$ as the temporal representation of node $i$ and update the memory $\mem^-$ by
$\mem^-_i \leftarrow \vemb_i(t-)$.

\subsection{Decoder}
\label{sec:method:dec}

In this paper, we consider the temporal link prediction task as it provides self-supervision signals.
Given $\vemb_i(t-)$ and $\vemb_j(t-)$ produced by the encoder, the decoder computes the probability of the event $e_{ij}(t)$ using a two-layer MLP followed by the sigmoid function $\sigma(\cdot)$.
\begin{equation}
\hat{p}_{i j}(t) = \sigma\bigg(\merge\big(\vemb_i(t-) \| \vemb_j(t-) \|\mathbb{I}\{i \in \bar{\hist}^n_j(t)\}\| \mathbb{I}\{j \in \bar{\hist}^n_i(t)\}  \big) \bigg),
\label{eq:bit}
\end{equation}
where $\mathbb{I}$ is the indicator function.
As observed in \cite{jodie,poursafaei2022towards}, there are many events reoccur over time in real-world temporal interaction graphs.
We introduce two extra bits, $\mathbb{I}\{i \in \bar{\hist}^n_j(t)\}$ and $\mathbb{I}\{j \in \bar{\hist}^n_i(t)\}$, to denote whether node $i$ recently interacted with node $j$.

Then we use the binary cross-entropy as the loss function:
\begin{equation}
\begin{aligned}
\mathcal{L}_1 = - \sum_{e_{ij}(t) \in \mathcal{E}} \big[\log \hat{p}_{ij}(t) + \log (1 - \hat{p}_{ik}(t)) \big],
\end{aligned}
\label{eq:link}
\end{equation}
where $k$ is randomly sampled as a negative destination node, and $\hat{p}_{ik}(t)$ is computed similarly with $\vemb_i(t-)$ and $\vemb_k(t-)$.

The encoder and the decoder constitute a complete temporal graph neural network, and we name this encoder-decoder model TIGE.
In Figure \ref{fig:dataflow:tige} we illustrate TIGE with $\mem^{\text{msg}} = \mem^-$ and $\mem^{\text{upd}} = \mem^+$.
By comparing data flows of TGN \cite{rossi2020temporal} and TIGE, we can observe that TGN roughly is a special case of TIGE with $\mem^{\text{msg}} = \mem^{\text{upd}} = \mem^+$.
Obviously, the most time-consuming module is the temporal embedding module.
However, TGN does not cherish its outputs and discards them after computing the loss.
In TIGE, the outputs of the embedding module are recycled to update the memory which then in turn makes the messages (Equation \ref{eq:msg_gen}) more effective.
We prove the necessity of introducing dual memory in the ablation study.

\subsection{Restarter}

At the given time $t$, the restarter aims to efficiently generate estimates of $\vemb(t-)$ and $\vemb(t+)$ using only a small portion of data $\bar{\mathcal{E}}(t) \subset \mathcal{E}(t)$, \ie, $|\bar{\mathcal{E}}| \ll |\mathcal{E}|$.
The outputs of the restarter, $\hat{\vemb}(t+)$ and $\hat{\vemb}(t-)$ can then be used as warm initialization of the memory.
To achieve this goal, we make the restarter mimic the encoder by minimizing the following distillation loss
\begin{equation}
\begin{aligned}
\mathcal{L}_2 = \sum_{e_{ij}(t) \in \mathcal{E}} \bigg[ & \quad \big\| \vemb_i(t+) - \hat{\vemb}_i(t+) \big\|_2^2 + \big\| \vemb_j(t+) - \hat{\vemb}_j(t+) \big\|_2^2 \\
& + \big\| \vemb_i(t-) - \hat{\vemb}_i(t-) \big\|_2^2 + \big\| \vemb_j(t-) - \hat{\vemb}_j(t-) \big\|_2^2  \bigg],
\end{aligned}
\label{eq:mse}
\end{equation}
where $\| \cdot \|_2^2$ denotes the squared $\ell_2$-norm, and $\vemb_{\cdot}(t+)$ and $\vemb_{\cdot}(t-)$ are computed by Equation \ref{eq:upd} and Equation \ref{eq:tgat} respectively.
This process is similar to knowledge distillation \cite{hinton2015distilling} where we transfer knowledge from the encoder to the student restarter model.

The restarter can take any form as long as it is efficient and does not require full data.
In this paper, we consider the following two forms of restarter.

\subsubsection{Transformer Restarter}
\label{sec:method:trans}

For a given event $e_{ij}(t)$, the Transformer restarter only takes node $i$'s ($j$'s) history sequence as input, and we have $\bar{\mathcal{E}}(t) = \cup_i \bar{\hist}_i^{m}(t)$.

Firstly, let us prepare the inputs of the Transformer.
Assume the most recent $m-1$ events involving node $i$ before time $t$ are
$e_{\pi(m-1)}(t_{\pi(m-1)}), \dots, e_{\pi(1)}(t_{\pi(1)})$,
where $t_{\pi(m-1)} < \dots < t_{\pi(1)} < t$.
To better capture the inductive evolving patterns especially for non-attributed graphs, we re-index $\{\pi(m-1), \dots, \pi(1), j\}$ and pass the new indices to a embedding lookup table to get $d$-dimensional vectors.
Since we limit the length of history, the embedding table can be represented as a $m \times d$ matrix.
We denote this process as the function $\reindex$.

Then for each event $e_{\pi(k)}(t_k)$, we use the following formula to compute its augmented representation:
\begin{equation}
\vec{x}_{k} = \vnode_{i} \| \vnode_{\pi(k)} \| \reindex(\pi(k)) \| \vedge_{\pi(k)}(t) \| \Phi(t-t_{\pi(k)}).
\label{eq:transformer_edge}
\end{equation}
Here we have $\vec{x}_{\pi(k)} \in \Re^{5d}$.
We assume node $i$ is the source node. Otherwise, we can swap the first two vectors in Equation \ref{eq:transformer_edge}.
Similarly, we encode the current event $e_{ij}(t)$ as 
\begin{equation*}
\begin{aligned}
{\vec{x}}_0 &=& \vnode_{i} &\|& \vnode_{j} &\|& \reindex(j) &\|& \vedge_{ij}(t) &\|& \Phi(0), \\
{\vec{x}}'_0 &=& \vec{0} &\|& \vec{0} &\|& \vec{0} &\|& \vec{0} &\|& \Phi(0). \\
\end{aligned}
\end{equation*}

Clearly, ${\vec{x}}'_0$ only contains the time information and is suitable for computing the representation of node $i$ just before time $t$:
\begin{equation*}
\hat{\vemb}_i(t-) = \transformer({\vec{x}}'_0, {\vec{x}}_1, \dots, {\vec{x}}_h ),
\end{equation*}
where $\transformer$ is a standard Transformer network \cite{vaswani2017attention} followed by a linear mapping to reduce the dimension from $5d$ to $d$.
Then we use another two-layer MLP to simulate the jump at $t$:
$\hat{\vemb}_i(t+) = \merge\bigl( \hat{\vemb}_i(t-) \| {\vec{x}}_0\bigr)$.

\subsubsection{Static Restarter}
\label{sec:method:static}

We also propose an extremely simple static restarter where $\bar{\mathcal{E}}(t)$ is empty.
For a static restarter, we maintain two embedding tables $\mathtt{static\_emb}^+$ and $\mathtt{static\_emb}^-$ initialized with zeros.
Then for any time $t$ we have 
$\hat{\vemb}_i(t+) =  \mathtt{static\_emb}^+(i)$ and 
$\hat{\vemb}_i(t-) =  \mathtt{static\_emb}^-(i)$.
If a new node is encountered during inference, the static restarter simply returns the zero vector.

Despite its simplicity, the static restarter requires $O(|\mathcal{V}|)$ parameters, which is contrast to the constant number of parameters of Transformer restarters.

\begin{table*}[t]
\centering
\caption{Average Precision (\%) for future edge prediction task in transductive and inductive settings. Best results are  highlighted in bold, and the best baselines are underlined.}
\begin{tabular}{c|c|cccc|ccc}
\toprule
\multicolumn{2}{c}{} & JODIE & TGAT & DyRep & TGN & TIGE & TIGER-T & TIGER-S \\
\midrule
\multirow{5}{*}{\rotatebox[origin=c]{90}{Transductive}}
 & Wikipedia  & 94.62 \tpm 0.5 & 95.34 \tpm 0.1 & 94.59 \tpm 0.2 & \underline{98.46 \tpm 0.1} & 98.83 \tpm 0.1 & \textbf{98.90 \tpm 0.0} & 98.81 \tpm 0.0 \\
 & Reddit     & 97.11 \tpm 0.3 & 98.12 \tpm 0.2 & 97.98 \tpm 0.1 & \underline{98.70 \tpm 0.1} & \textbf{99.04 \tpm 0.0} & 99.02 \tpm 0.0 & 98.96 \tpm 0.0 \\
 & MOOC       & 76.50 \tpm 1.8 & 60.97 \tpm 0.3 & 75.37 \tpm 1.7 & \underline{85.88 \tpm 3.0} & 89.64 \tpm 0.9 & 86.99 \tpm 1.6 & \textbf{89.96 \tpm 0.8} \\
 & LastFM     & 68.77 \tpm 3.0 & 53.36 \tpm 0.1 & 68.77 \tpm 2.1 & \underline{71.76 \tpm 5.3} & 87.85 \tpm 0.9 & 85.17 \tpm 0.2 & \textbf{89.38 \tpm 0.9} \\
 & Industrial & 95.46 \tpm 0.1 & 93.99 \tpm 0.2 & 95.33 \tpm 0.2 & \underline{96.68 \tpm 0.2} & \textbf{97.76 \tpm 0.1} & 97.14 \tpm 0.1 & 97.35 \tpm 0.1 \\
\midrule
\multirow{5}{*}{\rotatebox[origin=c]{90}{Inductive}} 
 & Wikipedia  & 93.11 \tpm 0.4 & 93.99 \tpm 0.3 & 92.05 \tpm 0.3 & \underline{97.81 \tpm 0.1} & 98.45 \tpm 0.1 & \textbf{98.58 \tpm 0.0} & 98.39 \tpm 0.0 \\
 & Reddit     & 94.36 \tpm 1.1 & 96.62 \tpm 0.3 & 95.68 \tpm 0.2 & \underline{97.55 \tpm 0.1} & 98.39 \tpm 0.1 & \textbf{98.59 \tpm 0.0} & 97.68 \tpm 0.2 \\
 & MOOC       & 77.83 \tpm 2.1 & 63.50 \tpm 0.7 & 78.55 \tpm 1.1 & \underline{85.55 \tpm 2.9} & \textbf{89.51 \tpm 0.7} & 86.42 \tpm 1.7 & 88.49 \tpm 0.7 \\
 & LastFM     & \underline{82.55 \tpm 1.9} & 55.65 \tpm 0.2 & 81.33 \tpm 2.1 & 80.42 \tpm 4.9 & 90.14 \tpm 1.0 & 89.11 \tpm 0.3 & \textbf{90.93 \tpm 0.5} \\
 & Industrial & 93.04 \tpm 0.9 & 87.69 \tpm 1.4 & 91.77 \tpm 1.3 & \underline{94.49 \tpm 0.3} & 96.89 \tpm 0.0 & \textbf{96.98 \tpm 0.2} & 96.22 \tpm 0.1 \\
\bottomrule
\end{tabular}
\label{tab:exp_main}
\end{table*}

\begin{table*}[t]
\centering
\caption{AUROC (\%) for dynamic/static node classification task.}
\begin{tabular}{c|cccc|ccc}
\toprule
 & JODIE & TGAT & DyRep & TGN & TIGE & TIGER-T & TIGER-S \\
\midrule
Wikipedia  & 84.84 \tpm 1.2 & 83.69 \tpm 0.7 & 84.59 \tpm 2.2 & \underline{87.81 \tpm 0.3} & 86.92 \tpm 0.7 & 85.16 \tpm 1.0 &  85.52 \tpm 0.8\\
Reddit     & 61.83 \tpm 2.7 & 65.56 \tpm 0.7 & 62.91 \tpm 2.4 & \underline{67.06 \tpm 0.9} &\textbf{69.41 \tpm 1.3} & 68.69 \tpm 1.7 & 68.31 \tpm 2.2 \\
MOOC       & 66.87 \tpm 0.4 & 53.95 \tpm 0.2 & \underline{67.76 \tpm 0.5} & 59.54 \tpm 1.0 & \textbf{72.35 \tpm 2.3} & 71.00 \tpm 2.2 & 72.16 \tpm 1.3 \\
Industrial & 77.10 \tpm 0.5 & 76.85 \tpm 0.4 & \underline{77.82 \tpm 0.3} & 75.96 \tpm 0.4 & \textbf{80.03 \tpm 0.4} & 76.60 \tpm 0.3 & 77.60 \tpm 0.5 \\
\bottomrule
\end{tabular}
\label{tab:exp_node}
\end{table*}

\begin{table*}[t]
\centering
\caption{Average Precision (\%) for future edge prediction task in transductive and inductive settings. Note that only \underline{20\%} of training data are used.}
\begin{tabular}{c|c|cccc|cc}
\toprule
\multicolumn{2}{c}{} & JODIE & TGAT & DyRep & TGN & TIGER-T & TIGER-S \\
\midrule
\multirow{5}{*}{\rotatebox[origin=c]{90}{Transductive}}
    & Wikipedia  & 79.28 \tpm 4.2 & 86.97 \tpm 1.7 & 88.19 \tpm 1.0 & \underline{96.34 \tpm 0.2} & 98.32 \tpm 0.1 & \textbf{98.36 \tpm 0.0} \\
    & Reddit     & 92.39 \tpm 1.4 & 97.19 \tpm 0.0 & 96.82 \tpm 0.3 & \underline{97.63 \tpm 0.1} & 98.67 \tpm 0.1 & \textbf{98.71 \tpm 0.0} \\
    & MOOC       & 55.73 \tpm 2.2 & 54.65 \tpm 0.4 & \underline{73.13 \tpm 1.7} & 56.54 \tpm 0.5 & 80.31 \tpm 0.6 & \textbf{79.73 \tpm 0.9} \\
    & LastFM     & \underline{68.00 \tpm 0.7} & 54.96 \tpm 0.4 & 67.38 \tpm 1.1 & 66.54 \tpm 2.0 & 84.53 \tpm 0.4 & \textbf{85.25 \tpm 0.3} \\
    & Industrial & 94.00 \tpm 0.1 & 92.21 \tpm 0.0 & 94.60 \tpm 0.0 & \underline{95.22 \tpm 0.4} & \textbf{97.00 \tpm 0.1} & 96.99 \tpm 0.0 \\
\midrule
\multirow{5}{*}{\rotatebox[origin=c]{90}{Inductive}} 
    & Wikipedia  & 79.30 \tpm 4.8 & 87.33 \tpm 1.1 & 85.99 \tpm 0.9 & \underline{95.86 \tpm 0.3} & 98.10 \tpm 0.1 & \textbf{98.17 \tpm 0.0} \\
    & Reddit     & 80.58 \tpm 2.8 & 95.18 \tpm 0.1 & 92.01 \tpm 0.8 & \underline{95.98 \tpm 0.4} & \textbf{98.12 \tpm 0.2} & 98.11 \tpm 0.1 \\
    & MOOC       & 58.51 \tpm 2.6 & 54.27 \tpm 1.0 & \underline{71.91 \tpm 2.1} & 61.11 \tpm 0.9 & 78.07 \tpm 0.5 & \textbf{77.41 \tpm 0.8} \\
    & LastFM     & \underline{80.96 \tpm 1.3} & 58.31 \tpm 0.7 & 79.67 \tpm 1.8 & 75.09 \tpm 2.8 & 88.54 \tpm 0.5 & \textbf{89.43 \tpm 0.2} \\
    & Industrial & \underline{93.64 \tpm 0.3} & 85.45 \tpm 0.0 & 92.94 \tpm 0.0 & 92.85 \tpm 1.3 & \textbf{96.75 \tpm 0.2} & 95.80 \tpm 0.7 \\
\bottomrule
\end{tabular}
\label{tab:exp_subset20}
\end{table*}

\subsection{Training and Inference}

With the help of the restarter, we can re-initialize the memory warmly at any time such that we can resume training/inference even if our model has been offline for a while, as illustrated in the Introduction.

Algorithm \ref{alg:single} describes how we train TIGER for one epoch.
The batch processing and the message aggregation are omitted for simplicity.
Since the restarter only uses limited data and cannot perfectly estimate memory, there will be some discrepancies.
To accustom the encoder to the surrogated memory, we occasionally re-initialize memory by the restarter during training, as shown in lines 5-8.
Specifically, at the beginning of each batch, we re-initialize the dual memory with probability $p_r > 0$.
Here the encoder and the restarter compose an interesting structure:
the encoder transfers its knowledge to the restarter by the distillation loss Equation \ref{eq:mse}, while the restarter conversely provides initial values to the encoder, as demonstrated in Figure \ref{fig:dataflow:tiger}.

Thanks to the restarter, we can also run TIGER in parallel by dividing the edge set $\mathcal{E}$ into multiple consecutive chunks as illustrated in Figure \ref{fig:butterfly:parallel}.
At the beginning of each chunk, we initialize the memory with the previous chunk.
We describe the multi-process version of TIGER in  Algorithm \ref{alg:parallel}.
All processes start from the same parameters (except memories), and gradients are synchronized in every backward pass.

The inference process of TIGER is similar to Algorithm \ref{alg:single} where the model parameters are fixed and we disable the restarter.

\section{Experiments}
\label{sec:exp}

\begin{figure}
\centering
\includegraphics[width=\linewidth]{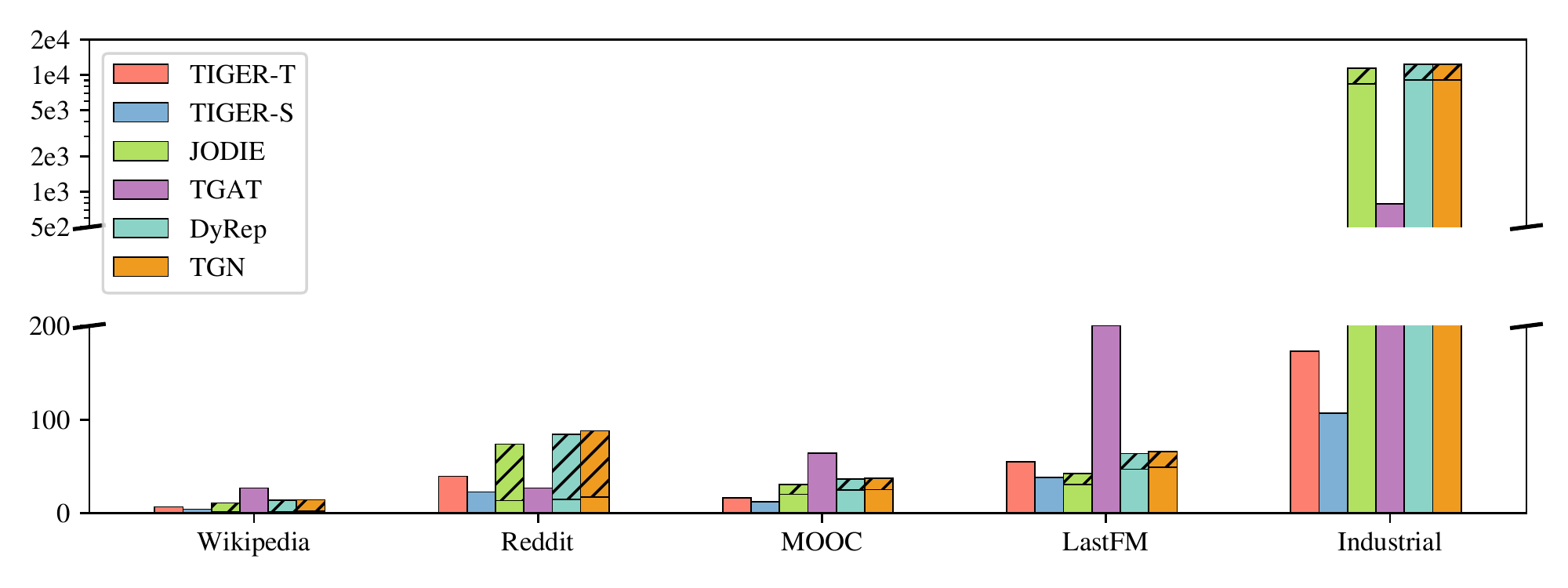}
\caption{Running times (per epoch in seconds) of different methods including training phases on 20\% of the training data and rollout phases on the remaining 80\%. The dashed areas denote the rollout phases.
Note that the upper part of the plot is of log scale.
}
\label{fig:subset20}
\Description{Running times of different methods.}
\end{figure}

\subsection{Datasets and Baselines}

We use four public datasets  \cite{jodie}, Wikipedia, Reddit, MOOC, LastFM, and one industrial dataset.
Specifically, the industrial dataset consists of one month of transactions sampled from a private real-world financial trading network from the Tencent Mobile Payment\footnote{The dataset is desensitized and sampled properly only for experiment purpose, and does not imply any commercial information. All personal identity information (PII) has been removed.}.
In this dataset, edges are with 170-d features, and nodes are with static labels indicating if they are fraudulent.
Details of public datasets are described in Appendix.
Note that all datasets are without node attributes, and MOOC and LastFM datasets are also without edge attributes.
Wikipedia, Reddit and LastFM are with dynamic labels indicating state changes of users.
For all datasets, we split edges chronologically by 70\%, 15\%, and 15\% for training, validation, and testing.
The statistics of the datasets are summarized in Appendix.

We choose JODIE \cite{jodie}, TGAT \cite{xu2020inductive}, DyRep \cite{trivedi2019dyrep}, TGN \cite{rossi2020temporal} as our baselines.
Static GNNs like GAT \cite{velivckovic2017graph} and methods that can only generate static representations like CTDNE \cite{ctdne} are not considered, as they have been shown inferior to the above models \cite{rossi2020temporal,jodie,poursafaei2022towards}.

As for our proposed methods, we consider TIGE, the basic encoder-decoder model introduced in Section \ref{sec:method:enc} and Section \ref{sec:method:dec}, and the two variants of TIGER, TIGER-T with the Transformer restarter (Section \ref{sec:method:trans}) and TIGER-S with the static restarter (Section \ref{sec:method:static}).
More implementation details and default hyper-parameters are discussed in Appendix.

\subsection{Temporal Link Prediction}
\label{sec:exp:main}

In the first group of experiments, we exactly follow the previous setting \cite{rossi2020temporal,xu2020inductive} to test the models' performance on \textbf{transductive} and \textbf{inductive} temporal link prediction tasks.
In the transductive setting, we examine edges whose nodes have been seen in the training splits.
On the contrary, we predict the temporal links between unseen nodes in the inductive setting.
We compute the average precision (AP) scores as the evaluation metric by sampling an equal amount of negative node pairs as we did in Equation \ref{eq:link}.

We show the results in Table \ref{tab:exp_main}.
Clearly, we can observe that all our proposed methods, TIGE, TIGER-T and TIGER-S, outperform the baselines in both transductive and inductive settings on all datasets.
This observation proves the effectiveness of introducing the dual memory.
Despite the possible information loss when substituting the memory with the restarters, TIGER-T/S still perform well.
Especially on the non-attributed datasets MOOC and LastFM, TIGER-S outperforms TIGE.
This may be owing to the static embeddings providing extra information about nodes.
Though TIGER-T is not as good as TIGE or TIGER-S, it still outperforms all the baselines. 
Note that in non-attributed graphs, TIGER-T only models the re-occurrence of nodes in Equation \ref{eq:transformer_edge}, which means re-occurrence patterns are an important feature for temporal interaction graphs.
TGAT, as a memory-less method, performs worst on these two datasets.

\subsection{Node Classification}
\label{sec:exp:node}

We consider node classification as a downstream task to verify the effectiveness of the learned temporal representations \cite{rossi2020temporal}.
We conduct dynamic node classification on Wikipedia, Reddit, and MOOC.
The goal is to predict whether a node would change its state, \eg, banned or drop-out, at some point.
Following \cite{rossi2020temporal,xu2020inductive}, we pass the time-aware representations $\vemb(t-)$ through a two-layer MLP to get the probabilities of state changes.
On the Industrial dataset where node labels are static, we compute the average representations of node $i$ at timestamps when its related events happen as the input of the MLP.

Our proposed methods achieve the best or comparable performance as shown in Table \ref{tab:exp_node}.
On all datasets, TIGE outperforms TIGER-T/S, indicating that the restarter may lead to relatively fluctuating temporal representations. 
Nonetheless, the satisfying results prove that the learned representations of our methods are effective for downstream tasks.

\subsection{Link Prediction with Limited Training Data}
\label{sec:exp:subset20}

Different from the previous experiments, in this group of experiments, we only use 20\% of the training data, \ie, 14\% of the total data, to simulate the scenarios discussed in the Introduction.

For JODIE, DyRep, and TGN, we first use 20\% of the training data to train the models and then update their memories on the remaining 80\% without computing the temporal embeddings and losses.
Take TGN as an example. The data flow from $\vec{s}(t')$ to $\mathcal{L}$ shown in Figure \ref{fig:dataflow:tgn} is disabled during this rollout phase.
TGAT only uses 20\% of the training data and discards the remaining parts as it is memory-less.
For TIGER-T and TIGER-S, we use 20\% of the data to train and then re-initialize the memory at the beginning of the validation split.
Note that this experiment group has the same validation/test data, including negative samples, as Section \ref{sec:exp:main}.

The results in Table \ref{tab:exp_subset20} clearly demonstrate the effectiveness of our proposed methods.
TIGER-T and TIGER-S trained with 20\% of data not only outperform all the baselines under the same setting, but also show superior performance to the baselines trained with 100\% of data on LastFM and Industrial datasets.

We also show the running times of the methods in Figure \ref{fig:subset20}.
JODIE, DyRep, and TGN have to keep running after the first 20\% of data to ensure their memories are up-to-date, which costs a lot.
TGAT is slow in large datasets due to its uniform neighbor sampling procedures \cite{rossi2020temporal}.
Thanks to the restarter modules in our TIGER, we can train the models efficiently and deploy them flexibly.

\subsection{Ablation Study}

\begin{table}[!t]
\caption{Ablation study.}
\label{tab:mem}
\begin{tabular}{c|cc|cc}
\toprule
\multicolumn{3}{c}{} & Wikipedia  & MOOC \\
\multicolumn{1}{c}{} & $\mem^{\text{msg}}$ & \multicolumn{1}{c}{$\mem^{\text{upd}}$} & &\\
\midrule
\multirow{5}{*}{\rotatebox[origin=c]{90}{TIGE}}
& \multicolumn{2}{c|}{TGN}
        & 98.46 \tpm 0.1 & 85.88 \tpm 3.0 \\
& + & + & 98.75 \tpm 0.0 & 88.28 \tpm 1.9 \\
& + & - & 98.77 \tpm 0.1 & 86.66 \tpm 1.5 \\
& - & - & 98.77 \tpm 0.1 & 86.50 \tpm 1.9 \\
& - & + & \textbf{98.83 \tpm 0.1} & \textbf{89.64} \tpm 0.9 \\
\midrule
\multirow{4}{*}{\rotatebox[origin=c]{90}{TIGER-S}}
& + & + & 98.73 \tpm 0.0 & 87.99 \tpm 0.9 \\
& + & - & 98.78 \tpm 0.0 & 86.41 \tpm 2.0 \\
& - & - & 98.80 \tpm 0.1 & 86.94 \tpm 2.4 \\
& - & + & \textbf{98.81 \tpm 0.0} & \textbf{89.96} \tpm 0.8 \\
\bottomrule
\end{tabular}
\end{table}

\begin{figure}
\begin{tabular}{@{}c@{}c@{}}
\begin{subfigure}{.5\linewidth}
\centering
\includegraphics[width=\linewidth]{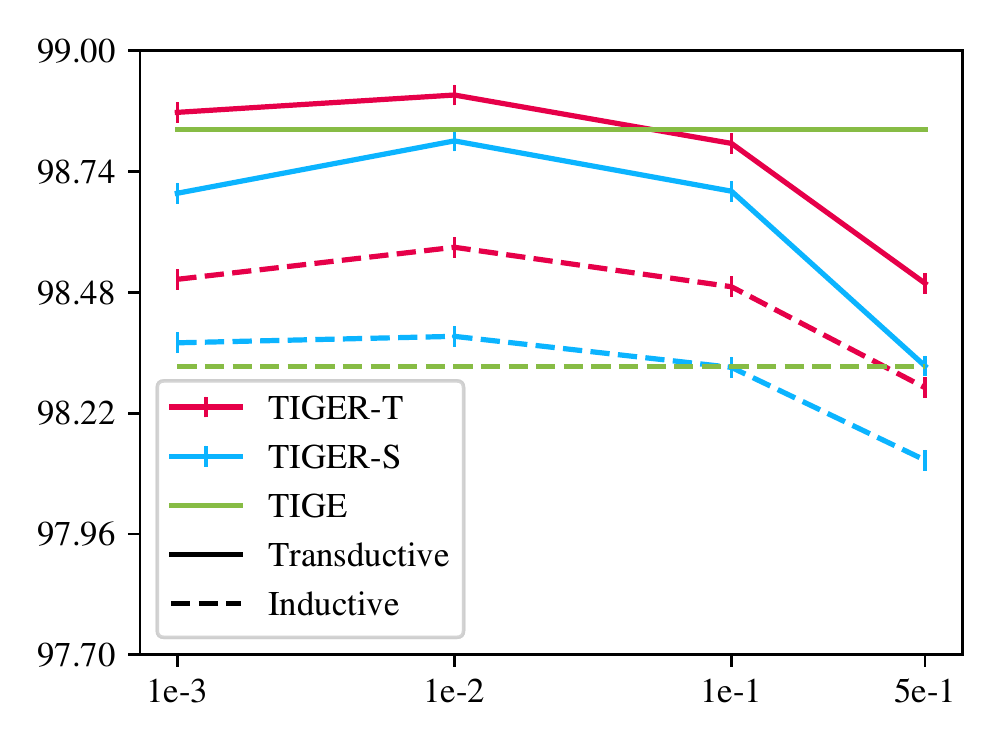}
\caption{Wikipedia}
\end{subfigure} &
\begin{subfigure}{.5\linewidth}
\centering
\includegraphics[width=\linewidth]{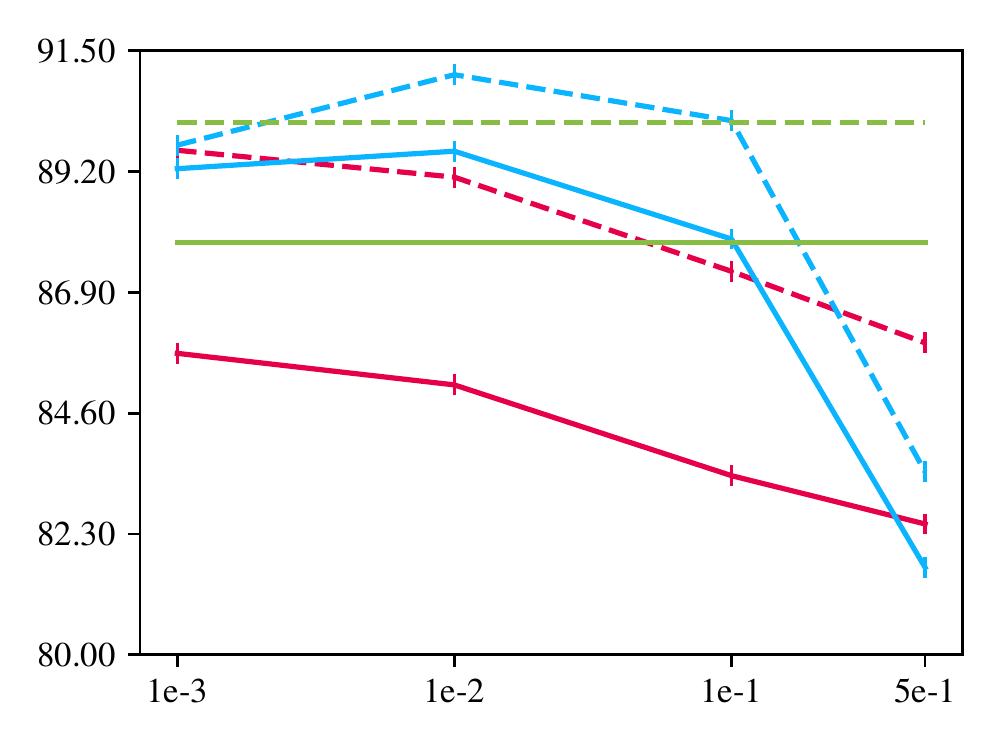}
\caption{LastFM}
\end{subfigure} 
\end{tabular}
\caption{Performance of models trained with different restart probabilities.
TIGE is shown for reference only.
}
\label{fig:prob}
\Description{Performance of models trained with different restart probabilities.}
\end{figure}

\begin{figure}
\begin{tabular}{@{}c@{}c@{}}
\begin{subfigure}{.5\linewidth}
\centering
\includegraphics[width=\linewidth]{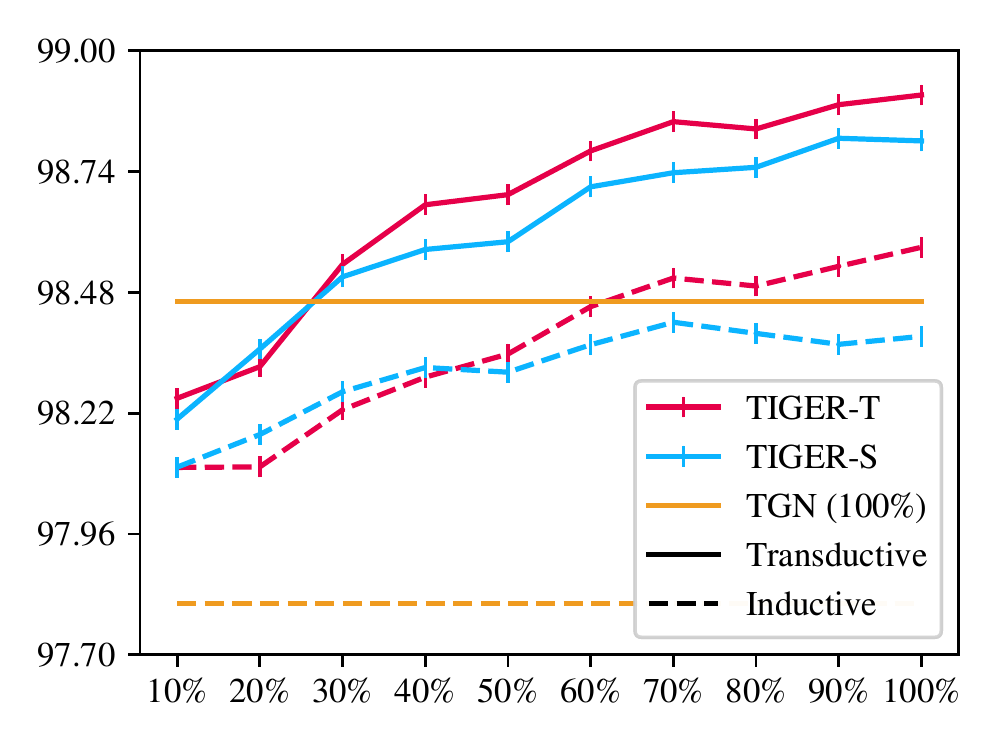}
\caption{Wikipedia}
\end{subfigure} &
\begin{subfigure}{.5\linewidth}
\centering
\includegraphics[width=\linewidth]{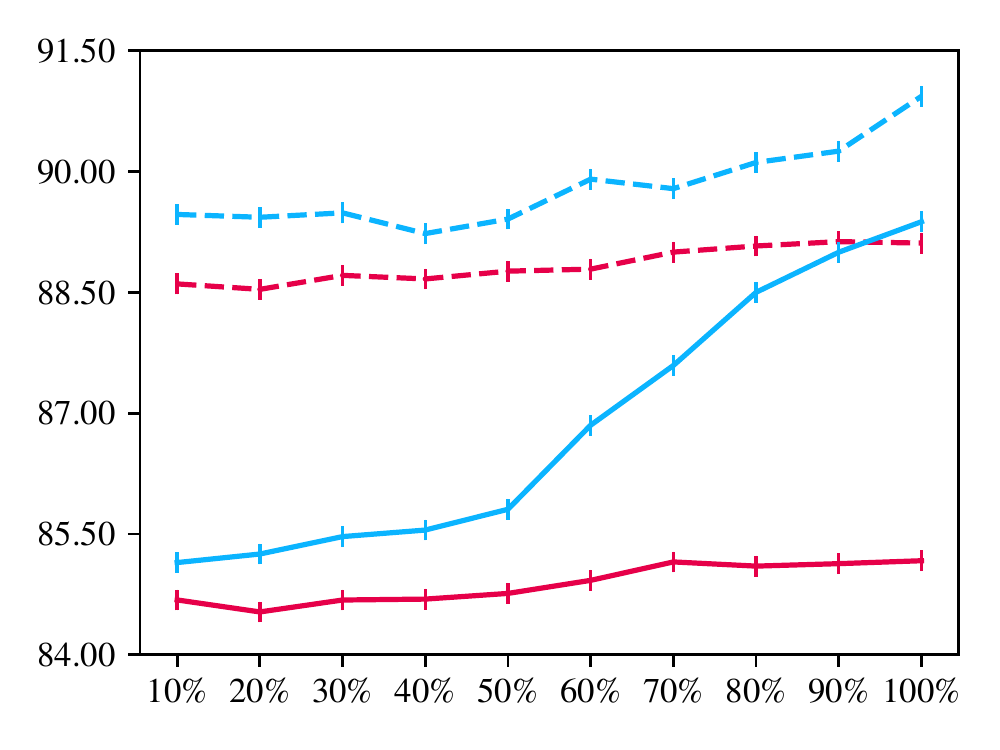}
\caption{LastFM}
\end{subfigure} 
\end{tabular}
\caption{Performance of models trained with different proportions of data.
TGN trained with 100\% of data is shown for reference (except on LastFM due to its poor performance).
}
\label{fig:subsets}
\Description{Performance of models trained with different proportions of data.}
\end{figure}

\begin{table}
\caption{Running times (per epoch in seconds) of TIGER-S w.r.t. the number of GPUs.}
\label{tab:gpu}
\begin{tabular}{c|rrr}
\toprule
\# GPU & 1 & 2 & 4 \\
\midrule
Wikipedia  &  30.3 &  17.1 (1.77x) &   8.9 (3.40x) \\
Reddit     & 181.6 & 102.0 (1.78x) &  51.9 (3.50x) \\
MOOC       &  47.3 &  34.2 (1.39x) &  21.5 (2.20x) \\
LastFM     & 190.1 & 139.2 (1.37x) &  76.2 (2.49x) \\
Industrial & 915.6 & 416.7 (2.20x) & 231.1 (3.96x) \\
\midrule
Avg. Speedup & 1x & 1.70x~ & 3.11x~ \\
\bottomrule
\end{tabular}
\end{table}

Recall that in our proposed methods we have $\mem^{\text{msg}} \in \{\mem^+, \mem^- \}$ and $\mem^{\text{upd}} \in \{\mem^+, \mem^- \}$, leading to 4 combinations in total.
In Table \ref{tab:mem} we present the results of transductive link prediction under different combinations on two representative datasets. The results of TIGER-T are shown in Appendix due to page limit.

As we illustrated in Figure \ref{fig:dataflow}, TGN can be seen as a special case of TIGE with $(+, +)$ memory.
However, in TIGE we further introduce two bits of information (Equation \ref{eq:bit}) to capture re-occurrence patterns.
The first two lines in Table \ref{tab:mem} prove the necessity of this module.
Unfortunately, we find that $\mem^{\text{upd}}=\mem^-$ would result in relatively poor performance (but still better than baselines).
We infer that this is because the GRU-based update function is more compatible with recurrent updates, \eg, $\vemb(t'+) \rightarrow \vemb(t+)$, rather than simulating jumps, \eg, $\vemb(t-) \rightarrow \vemb(t+)$ \cite{zhang2021cope}.
All of our proposed methods achieve the best results with the default combination $(-, +)$.
This is because $\vemb(t-)$ also encodes the temporal neighborhood information (Equation \ref{eq:tgat}), which makes it more expressive when generating messages.

\subsection{Parameter Study}

To accustom the encoder to the surrogated memory, we occasionally re-initialize memory by the restarter during training.
Now we study how the frequency of restarts affects the performance.
We plot APs of TIGER-T/S with varying restart probabilities $p_r$ in Figure \ref{fig:prob}. 
Results on Reddit and MOOC are shown in Appendix due to the page limit.
We can observe that restarting too often ($p_r \ge 0.1$) may make models hard to learn long-term dependencies and thus result in degenerated performance, while restarting too less often ($p_r = 0.001$) may cause the encoder failing to accustom to the estimated values.
And $p_r=0.01$ seems to be a sweet spot.

\subsection{Performance w.r.t the Proportion of Training Data} 

We extend the setting in Section \ref{sec:exp:subset20}.
We plot the results of TIGER-T/S by varying the proportion of training data from 10\% to 100\% in Figure \ref{fig:subsets}.
Results on Reddit and MOOC are shown in Appendix due to the page limit.
We find that our proposed methods constantly improve with more training data.
It is also worth noting that in the inductive setting on Wikipedia, Reddit, and LastFM, our proposed models trained with only 10\% of data can beat the best baseline TGN trained with full data.
This also proves the generalizability of our methods.

\subsection{Multi-GPU Parallel}

As a direct product of the restarters, our proposed TIGER can easily be trained in parallel as shown in Algorithm \ref{alg:parallel}.
The running times of TIGER-S under multiple GPUs are shown in Table \ref{tab:gpu},
and we can observe that the multi-GPU parallelizing brings satisfying speedups especially on the largest Industrial dataset.
This demonstrates that our proposed method TIGER is highly scalable.

\section{Conclusion}
\label{sec:conclusion}

In this paper, we propose TIGER, a temporal graph neural network that can flexibly restart at any time. 
By introducing the dual memory, we further alleviate the staleness problem and improve the interpretability of memory.
The restarter module then enables our models to restart at any time and run in parallel.

As for future works, we will consider other types of restarters to fit different datasets.
For example, the memory-less model TGAT seems a possible candidate.
In this paper, we parallelize our methods by splitting the edges into different chunks.
It also seems possible to split the graph into several subgraphs with the proposed restarters.
This may help the methods scale to graphs with large node sets.

\begin{acks}
This work is funded in part by the National Natural Science Foundation of China Projects No. 62206059, No. U1936213 and China Postdoctoral Science Foundation 2022M710747.
\end{acks}

\bibliographystyle{ACM-Reference-Format}
\bibliography{ref}

\pagebreak
\appendix
\section{Appendix}

The codes of our methods are available at \url{https://github.com/yzhang1918/www2023tiger}.

\subsection{Datasets}

\begin{table}[h]
\centering
\caption{Statistics of datasets.}
\label{tab:data}
\begin{tabular}{c|cccc}
\toprule
& \# Nodes & \# Edges & \# Edge Features & \# Labels \\
\midrule
Wikipedia & 9,227 & 157,474 & 172 & Dynamic \\
Reddit & 10,984 & 672,447 & 172 & Dynamic \\
MOOC & 7,144 & 411,749 & 0 & Dynamic \\
LastFM & 1,980 & 1,293,103 & 0 & - \\
Industrial & 188,538 & 3,840,240 & 170 & Static \\
\toprule
\end{tabular}
\end{table}

We use the following public datasets\footnote{\url{https://snap.stanford.edu/jodie/}} provided by the authors of JODIE \cite{jodie}. 
(1) Wikipedia dataset contains edits of Wikipedia pages by users.
(2) Reddit dataset consists of users' posts on subreddits. 
In these two datasets, edges are with 172-d feature vectors,
and user nodes are with dynamic labels indicating if they get banned after some events.
(3) MOOC dataset consists of actions done by students on online courses, and nodes are with dynamic labels indicating if students drop-out of courses.
(4) LastFM dataset consists of events that users listen to songs.
MOOC and LastFM datasets are non-attributed.
The statistics of the datasets are summarized in Table \ref{tab:data}.

\subsection{Implementation Details}

\begin{table}[h]
\centering
\caption{Default values of hyper-parameters in \ours. }
\label{tab:deafult}
\begin{tabular}{c|c}
\toprule
Hyper-parameter & Value \\
\midrule
(TIGE) & \\
Number of GNN layers $L$ & 1 \\
Number of attention heads & 2 \\
Number of GNN neighbors $n$ & 10 \\
Batch size & 200 \\
Learning rate & 0.0001 \\
Optimizer & Adam \\ 
Dropout rate & 0.1 \\
Restart probability $p_r$ & 0.01\\
\midrule
(Transformer Restarter) & \\
Number of layers & 1 \\
Number of heads & 2 \\
Length of history $m$ & 40 \\
\bottomrule
\end{tabular}
\end{table}

We implement our methods in PyTorch\footnote{\url{https://pytorch.org/}} based on the official implementation\footnote{\url{https://github.com/twitter-research/tgn/}} of TGN \cite{rossi2020temporal}.

Unless otherwise stated, we use the default hyper-parameters summarized in Table \ref{tab:deafult}.
Note that except the restarter-related parameters, we use the same hyper-parameters as TGN for a fair comparison.
For attributed datasets, Wikipedia, Reddit, and Industrial, the model dimension is equal to the dimension of their edge features.
For non-attributed graphs, the dimension is set as 100 for all methods.

Since we strictly follow the settings in TGN \cite{rossi2020temporal}, we reuse the results of temporal link prediction (Section \ref{sec:exp:main}) and node classification (Section \ref{sec:exp:node}) on Wikipedia and Reddit reported in \cite{rossi2020temporal}.

Experiments on public datasets were conducted on a single server with 72 cores, 128GB memory, and four Nvidia Tesla V100 GPUs.

\subsection{Pseudo-codes}

The pseudo-codes of our methods are shown in Algorithm \ref{alg:single} and \ref{alg:parallel}.

\begin{algorithm}[!htp]
\caption{TIGER (one epoch, single-process)}
\label{alg:single}
\begin{algorithmic}[1]
\Require Edge set $\mathcal{E}$, (optional) memory $\mem^+$, $\mem^-$
\If{ $\mem^+, \mem^-$ not provided}
\State Initialize $\mem^+, \mem^- \leftarrow \vec{0}$ 
\EndIf
\For{$e_{ij}(t) \in \mathcal{E} $}
\State Sample $p \sim \mathtt{uniform}[0, 1]$
\If{$p < p_r$}  \Comment{restart}
    \State Re-initialize $\mem^+, \mem^- \leftarrow \mathtt{restarter}(\mathcal{E}(t'))$
\EndIf
\State Load $\vemb(t''+) \leftarrow \mem^+ , ~ \vemb(t'-) \leftarrow \mem^-$
\State Compute $\vemb(t'+), \vemb(t-) = \mathtt{encoder}(\vemb(t''+), \vemb(t'-))$
\State Update $\mem^+ \leftarrow \vemb(t'+), ~ \mem^- \leftarrow \vemb(t-)$
\State Compute $\hat{p}_{ij}(t), \hat{p}_{ik}(t) = \mathtt{decoder}(\vemb(t-))$
\State Compute temporal link prediction loss $\mathcal{L}_1 = -\log \hat{p}_{ij}(t) - \log(1- \hat{p}_{ik}(t))$ and update parameters of $\mathtt{encoder}, \mathtt{decoder}$
\State Compute  $\hat{\vemb}(t'-), \hat{\vemb}(t'+) = \mathtt{restarter}(\mathcal{E}(t'))$
\State Compute distillation loss  $\mathcal{L}_2=\|\vemb(t'-) - \hat{\vemb}(t'-)\|_2^2+\|\vemb(t'+) - \hat{\vemb}(t'+)\|_2^2$ and update parameters of $\mathtt{restarter}$
\EndFor
\end{algorithmic}
\end{algorithm}

\begin{algorithm}[!htp]
\caption{TIGER (one epoch, multi-process)}
\label{alg:parallel}
\begin{algorithmic}[1]
\Require Edge set $\mathcal{E}$, number of processes $N$
\State Divide $\mathcal{E}$ equally into $\mathcal{E}_1, \dots \mathcal{E}_N$
\State Initialize $\mem_1^+, \mem_1^- \leftarrow \vec{0}$
\For{$k = 2, \dots, N$} \Comment{restart}
    \State Initialize  $\mem_k^+, \mem_k^- \leftarrow \mathtt{restarter}(\mathcal{E}_{k-1})$
\EndFor
\For{$k = 1, \dots, N$} \Comment{run in parallel}
    \State Call Algorithm \ref{alg:single} with inputs $(\mathcal{E}_k, \mem_k^+, \mem_k^-)$
\EndFor
\end{algorithmic}
\end{algorithm}

\subsection{Supplementary Results}

\begin{figure}[h]
\begin{tabular}{@{}c@{}c@{}}
\begin{subfigure}{.5\linewidth}
\centering
\includegraphics[width=\linewidth]{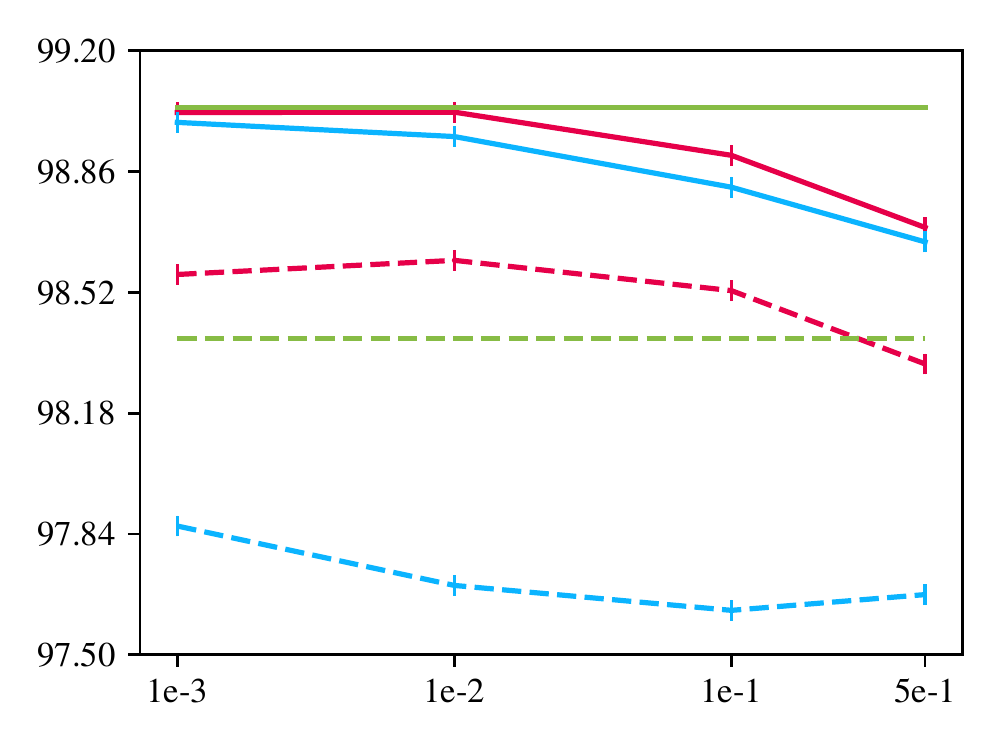}
\caption{Reddit}
\end{subfigure} &
\begin{subfigure}{.5\linewidth}
\centering
\includegraphics[width=\linewidth]{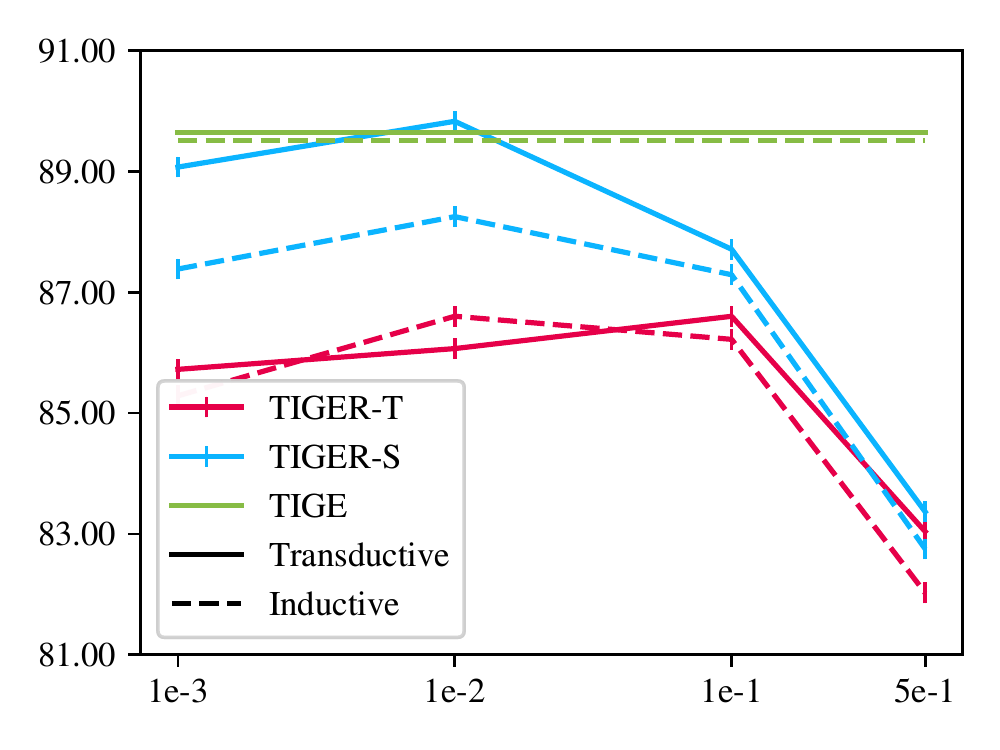}
\caption{MOOC}
\end{subfigure}
\end{tabular}
\caption{%
Performance of models trained with different restart probabilities.
TIGE is shown for reference only.
(Supplementary results for Figure \ref{fig:prob})
}
\label{fig:prob2}
\Description{Performance of models trained with different restart probabilities.}
\end{figure}

\begin{figure}[h]
\begin{tabular}{@{}c@{}c@{}}
\begin{subfigure}{.5\linewidth}
\centering
\includegraphics[width=\linewidth]{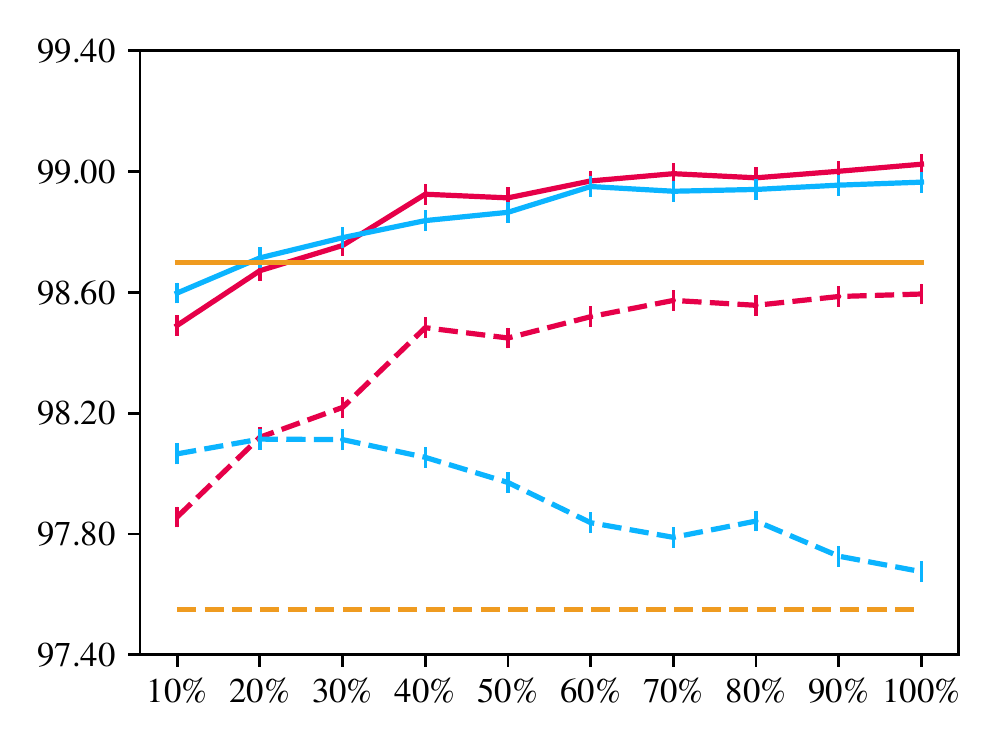}
\caption{Reddit}
\end{subfigure} &
\begin{subfigure}{.5\linewidth}
\centering
\includegraphics[width=\linewidth]{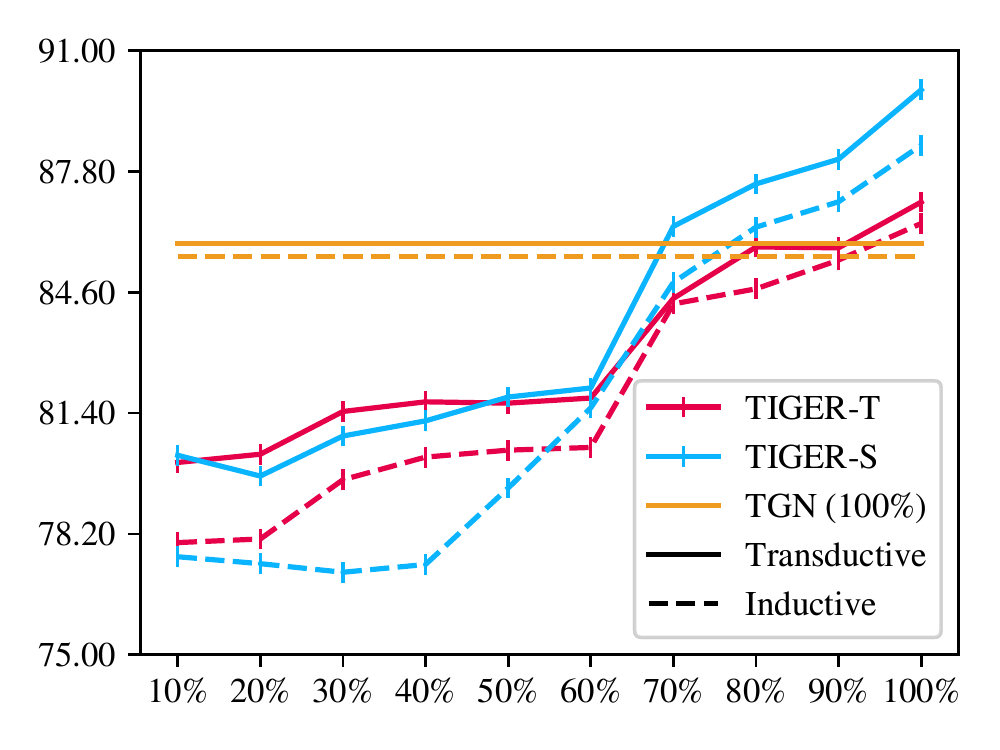}
\caption{MOOC}
\end{subfigure} 
\end{tabular}
\caption{%
Performance of models trained with different proportions of data.
TGN trained with 100\% of data is shown for reference.
(Supplementary results for Figure \ref{fig:subsets})
}
\label{fig:subsets2}
\Description{Performance of models trained with different proportions of data.}
\end{figure}

Due to the space limit, we show the supplementary results for Table \ref{tab:mem}, Figure \ref{fig:prob}, and Figure \ref{fig:subsets} in Table \ref{tab:mem2}, Figure \ref{fig:prob2}, and Figure \ref{fig:subsets2} respectively.

\begin{table}[!h]
\caption{Ablation study (Supplementary results for Table \ref{tab:mem})}
\label{tab:mem2}
\begin{tabular}{c|cc|cc}
\toprule
\multicolumn{3}{c}{} & Wikipedia  & MOOC \\
\multicolumn{1}{c}{} & $\mem^{\text{msg}}$ & \multicolumn{1}{c}{$\mem^{\text{upd}}$} & &\\
\midrule
\multirow{4}{*}{\rotatebox[origin=c]{90}{TIGER-T}}
& + & + & 98.80 \tpm 0.0 & 84.90 \tpm 2.3 \\
& + & - & 98.82 \tpm 0.1 & 83.81 \tpm 0.8 \\
& - & - & 98.85 \tpm 0.1 & 84.97 \tpm 2.5 \\
& - & + & \textbf{98.90 \tpm 0.0} & \textbf{86.99} \tpm 1.6 \\
\bottomrule
\end{tabular}
\end{table}

During the rebuttal phase, the reviewers suggested to include CAW-N \cite{wang2021inductive} and PINT \cite{souza2022provably} as baselines.
CAW-N examines joint structure information with causal anonymous walks.
CAW-N is typically not considered as a standard temporal GNN-based method \cite{souza2022provably}.
The authors of PINT theoretically proved the representational power and limits of current temporal GNNs and proposed a more expressive model with injective temporal message passing and relative positional features.
As the code of PINT had not been uploaded at the time we finished the camera-ready version of the paper, we reused their table numbers on the three common datasets.
From Table \ref{tab:pint} we can observe that \ours{} outperforms PINT and CAW-N in most cases.
As LastFM has no features, we show results of \ours{} with dim=172 to align with PINT's setting. (We used dim=100 in \ref{tab:exp_main}.) 
Moreover, PINT uses 2-hop information while \ours{} uses 1-hop information. Thus the performance of \ours{} could be further improved by letting the number of GNN layers $L=2$.
The main gain of PINT comes from the positional features, which are expensive to compute (please refer to Figure 7 in PINT \cite{souza2022provably}), while our modifications, \ie, dual memory and restarters, are relatively lightweight.
PINT focuses on the theoretical aspects of temporal GNNs, which is orthogonal to our work. It is interesting to see if our dual memory and restarter modules could further improve PINT.

\begin{table}[h]
\centering
\caption{Average Precision (\%) for future edge prediction task in transductive and inductive settings.}
\begin{tabular}{cc|ccc}
\toprule
&& Wikipedia & Reddit & LastFM* \\
\midrule
\multirow{5}{*}{\rotatebox[origin=c]{90}{Transductive}} 
& CAW-N   & 98.63 \tpm 0.1 & 98.39 \tpm 0.1 & 81.29 \tpm 0.1 \\
& PINT    & 98.78 \tpm 0.1 & 99.03 \tpm 0.0 & 88.06 \tpm 0.7 \\
& TIGE    & 98.83 \tpm 0.1 & \textbf{99.04 \tpm 0.0} & 89.18 \tpm 0.6 \\
& TIGER-T & \textbf{98.90 \tpm 0.0} & 99.02 \tpm 0.0 & 85.36 \tpm 0.3 \\
& TIGER-S & 98.81 \tpm 0.0 & 98.96 \tpm 0.0 & \textbf{90.31 \tpm 0.5} \\
\midrule
\multirow{5}{*}{\rotatebox[origin=c]{90}{Inductive}} 
& CAW-N   & 98.52 \tpm 0.1 & 97.81 \tpm 0.1 & 85.67 \tpm 0.5 \\
& PINT    & 98.38 \tpm 0.0 & 98.25 \tpm 0.0 & \textbf{91.76 \tpm 0.7} \\ 
& TIGE    & 98.45 \tpm 0.1 & 98.39 \tpm 0.1 & 91.01 \tpm 0.6 \\
& TIGER-T & \textbf{98.58 \tpm 0.0} & \textbf{98.59 \tpm 0.0} & 89.28 \tpm 0.4 \\
& TIGER-S & 98.39 \tpm 0.0 & 97.68 \tpm 0.2 & 91.00 \tpm 0.5 \\
\bottomrule
\end{tabular}
\label{tab:pint}
\end{table}

\end{document}